\documentclass[nonacm,sigconf]{aamas} 

\usepackage{balance} 

\doi{VRTD1803}

\makeatletter
\gdef\@copyrightpermission{
  \begin{minipage}{0.2\columnwidth}
   \href{https://creativecommons.org/licenses/by/4.0/}{\includegraphics[width=0.90\textwidth]{by}}
  \end{minipage}\hfill
  \begin{minipage}{0.8\columnwidth}
   \href{https://creativecommons.org/licenses/by/4.0/}{This work is licensed under a Creative Commons Attribution International 4.0 License.}
  \end{minipage}
  \vspace{5pt}
}
\makeatother

\setcopyright{ifaamas}
\acmConference[AAMAS '26]{Proc.\@ of the 25th International Conference
on Autonomous Agents and Multiagent Systems (AAMAS 2026)}{May 25 -- 29, 2026}
{Paphos, Cyprus}{C.~Amato, L.~Dennis, V.~Mascardi, J.~Thangarajah (eds.)}
\copyrightyear{2026}
\acmYear{2026}
\acmDOI{}
\acmPrice{}
\acmISBN{}

\usepackage{latexsym}
\usepackage{amsmath}
\usepackage{amsthm}
\usepackage{booktabs}
\usepackage{enumitem}
\usepackage{graphicx}
\usepackage{color}

\usepackage{forest}
\usepackage{wrapfig}
\usepackage{framed}
\usepackage{mdframed}
\usepackage{stmaryrd}
\usepackage{varwidth}
\usepackage{pifont}
\usepackage{multirow}
\usepackage{pdfpages}

\definecolor{darkgreen}{RGB}{0,149,48}
\definecolor{myorange}{RGB}{255,116,40}
\definecolor{mygreen}{RGB}{0,195,0}

\newcommand{\yes}{\textcolor{mygreen}{\ding{52}}}
\newcommand{\no}{\textcolor{red}{\ding{56}}}

\usepackage{listings}
\lstdefinestyle{mystyle}{
    backgroundcolor=\color{backcolour},   
    commentstyle=\color{codegreen},
    keywordstyle=\color{magenta},
    numberstyle=\tiny\color{codegray},
    stringstyle=\color{codepurple},
    basicstyle=\footnotesize,
    breakatwhitespace=true,         
    breaklines=true,                 
    captionpos=b,                    
    keepspaces=false,                 
    numbers=left,                    
    numbersep=4pt,                  
    showspaces=false,                
    showstringspaces=false,
    showtabs=false,                  
    tabsize=2,
}

\newcommand{\pre}{\mathit{pre}}
\newcommand{\cond}{\mathit{cond}}
\newcommand{\ancestor}{\mathit{ancest}}

\newcommand{\E}{\mathcal{E}}

\newcommand{\seqb}{\mathit{seq\_bef}}
\renewcommand{\seqb}{\mathit{seqBef}}

\newcommand{\BibTeX}{B\kern-.05em{\sc i\kern-.025em b}\kern-.08em\TeX}

\usepackage{dashbox}

\newcommand{\orig}[1]{\fcolorbox{black}{white}{$#1$}}
\newcommand{\cont}[1]{\fcolorbox{black}{lime}{$#1$}}
\newcommand{\conts}[1]{\fcolorbox{black}{lime}{$\scriptstyle #1$}}
\renewcommand{\orig}[1]{\dbox{$#1$}}

\setcopyright{ifaamas}
\acmConference[AAMAS '26]{Proc.\@ of the 25th International Conference
on Autonomous Agents and Multiagent Systems (AAMAS 2026)}{May 25 -- 29, 2026}
{Paphos, Cyprus}{C.~Amato, L.~Dennis, V.~Mascardi, J.~Thangarajah (eds.)}
\copyrightyear{2026}
\acmYear{2026}
\acmDOI{}
\acmPrice{}
\acmISBN{}

\acmSubmissionID{24} 

\title{Contrastive explanations of BDI agents }
\subtitle{AAMAS 2026 paper with additional supplementary material}

\author{Michael Winikoff}
\affiliation{
  \institution{Victoria University of Wellington}
  \city{Wellington}
  \country{New Zealand}}
\email{michael.winikoff@vuw.ac.nz}

\begin{abstract}
The ability of autonomous systems to provide explanations is important for supporting transparency and aiding the development of (appropriate) trust. Prior work has defined a mechanism for Belief-Desire-Intention (BDI) agents to be able to answer questions of the form ``why did you do action $X$?''. However, we know that we ask \emph{contrastive} questions (``why did you do $X$ \emph{instead of} $F$?''). We therefore extend previous work to be able to answer such questions. A computational evaluation shows that using contrastive questions yields a significant reduction in explanation length. A human subject evaluation was conducted to assess whether such contrastive answers are preferred, and how well they support trust development and transparency. We found some evidence for contrastive answers being preferred, and some evidence that they led to higher trust, perceived understanding, and confidence in the system's correctness. We also evaluated the benefit of providing explanations at all. Surprisingly, there was not a clear benefit, and in some situations we found evidence that providing a (full) explanation was worse than not providing any explanation. 
\end{abstract}


\newcommand{\supp}[1]{\cite[#1]{arXivSupp}}
\renewcommand{\supp}[1]{#1}

\begin{document}


\pagestyle{fancy}
\fancyhead{}


\maketitle 


\section{Introduction}

Explainability of autonomous systems~ (e.g.~\cite{commercial07,AppSurvey2014,DBLP:conf/atal/RiemsdijkJL15}) is important for a number of reasons. These include supporting transparency~\cite{DBLP:journals/firai/WinfieldBDEHJMO21,IEEEP7001,HLEGTrustworthyAssessment}, aiding the development of \emph{appropriate} levels of trust~\cite{DBLP:conf/aaai/LangleyMSC17,EMAS2017Winikoff,DBLP:conf/hri/RobinetteLAHW16}, and a range of other reasons (e.g.~acceptability~\cite{Floridi2018}, understandability~\cite{DBLP:conf/atal/VerhagenNT21}, accountability~\cite{DBLP:conf/at/CranefieldOV18}, and traceability~\cite{DBLP:conf/atal/VerhagenNT21}). 
In particular, it is important to provide means of engineering autonomous systems that can explain their behaviour in human-meaningful terms~\cite{DBLP:conf/atal/SRJT24b,DBLP:conf/atal/SRJT24a,DBLP:journals/ai/Miller19,DBLP:journals/ai/WinikoffSDD21}.

There is a whole body of work on explainable AI (XAI)~\cite{DBLP:conf/atal/AnjomshoaeNCF19}: techniques that allow explanations to be provided for the behaviour of AI modules. However, despite the importance of explainability of autonomous systems, most of the work on XAI has focused on explaining machine learning (``data-driven XAI''~\cite{DBLP:conf/atal/AnjomshoaeNCF19}), with a much smaller body of work  focusing on explaining autonomous agents (``goal-driven XAI''~\cite{DBLP:conf/atal/AnjomshoaeNCF19}, or ``explainable agency''~\cite{DBLP:conf/aaai/LangleyMSC17}). 

Prior work exploring how humans explain themselves~\cite{MalleBook} has shown that in similar contexts (i.e.~explaining the reasons for choosing a specific course of action) humans use the concepts of beliefs, desires, and \textit{valuings}. Subsequently, \citet{DBLP:journals/ai/WinikoffSDD21} noted the natural correspondence between these concepts and the Belief-Desire-Intention (BDI) agent architecture~\cite{KR92,BDI,Bratman}, and defined a mechanism\footnote{Earlier work by Harbers \textit{et al.}~\cite{Harbers2011,DBLP:conf/mates/BroekensHHBJM10,DBLP:conf/iat/HarbersBM10} also considered BDI agents and defined a mechanism for providing explanations. However, they defined particular fixed patterns for extracting reasons from goal-plan trees, rather than a general algorithm, and did not make the link to the evidence on how humans explain themselves.} that allows BDI agents 
to provide explanations of their actions in terms of these concepts,  answering questions of the form ``Why did you do action $X$?''. 

However, it is known that as humans we often ask  \emph{contrastive} questions of the form ``Why did you do $X$ (the \textit{fact}) \emph{instead of $F$} (the \textit{foil})?'' (although sometimes the ``instead of $F$'' is implicit)~\cite{DBLP:journals/ai/Miller19,DBLP:journals/jair/KrarupKMLC021}. 
There is also empirical evidence that such explanations are most effective~\cite{9575917}.

Although there has been some recent work on generating contrastive explanations\footnote{These are related but distinct from counterfactuals: a contrastive question is asking ``Why?'' about something that was actually done (with reference to a possible alternative), whereas a counterfactual question is a \emph{hypothetical} about something that was \emph{not} done.} in different settings (e.g.~planning~\cite{DBLP:journals/ai/SreedharanSK21,borgo2018providingexplanationsaiplanner,cashmore2019explainableaiplanningservice,10.1145/3561532}, rule-based systems~\cite{10.1145/3648505.3648507}, and Markov Decision Processes~\cite{DBLP:conf/aaai/AmitaiSA24,10.1109/IROS45743.2020.9341773}) and domains (e.g.~driving~\cite{BalintArXiv2206,9575917} and robotics~\cite{korpan2017whynaturalexplanationsrobot,olivaresalarcos2024contrastive}), we are not aware of any work that addresses generating contrastive explanations of actions for BDI agents. The closest is~\citet{10.1007/978-3-031-45368-7_23} which focuses on contrastive explanations for the \emph{selection} of goals, not the actions taken by the agent.

This paper makes three contributions: 
(1) extending \citet{DBLP:journals/ai/WinikoffSDD21} to generate contrastive explanations for BDI agents (\S\ref{sec:contrastive}); 
(2) conducting a \emph{computational} evaluation (\S\ref{sec:eval}) to assess the extent of the reduction in explanation length; and
(3) conducting a \emph{human subject} evaluation (\S\ref{sec:human_subject_evaluation}) to assess whether contrastive explanations are \emph{preferred} and whether they are \emph{effective}.

\section{Background}\label{sec:bg}

We firstly (\S\ref{sec31}) define goal-plan trees and then (\S\ref{sec32}) define (non-contrastive) explanation generation.

\subsection{Goal-Plan Trees}\label{sec31}

A goal-plan tree is a standard abstraction of a wide range of BDI agent platforms, which specify agent behaviour using event-triggered plans with a context condition and plan body. 
A goal-plan tree is either an \emph{action} node (which has no children), or a \emph{goal} node (which has at least one child with each child having a condition).  Where a node has multiple children we refer to them as ``siblings'', and to the ones appearing earlier in the sequence as being ``older''. All nodes have a \emph{name}.
Action nodes also have a pre- and post-condition. Goal nodes also have a \emph{type}. At a high level we have \textsc{And} (do all child nodes) and \textsc{Or} nodes (select one child node and do it). However, we also make a number of other distinctions, which lead us to use the following node types:  $all$ (all children are done in unconstrained order), 
$seq$ (all children are done in sequential order), $one$ (one child is selected and done), $sone$ (one child is selected and done, but the selection process considers the children in a specified order, i.e.~for a child to be selected its condition must be True and the conditions of all its ``older'' siblings must be False), and $xone$ (one child is selected and done, but, for explanatory purposes, it is indicated that the children are mutually exclusive: there is no situation in which more than one child is available for selection). 
We use \textsc{And} to refer to $all$ and $seq$ and \textsc{Or} to refer to the $one$, $sone$ and $xone$ node types. 
Formally we define a goal tree $GT$ as follows:
\begin{eqnarray*}
    GT & ::= & (ActionName,Pre,Post)  \; | \;  (GoalName,Type, Child^+) \\
    Child & ::= & (Cond,GT) \\
    Type & ::= & all \;|\; seq \;|\; one \;|\; sone \;|\; xone 
\end{eqnarray*}

We use as an example the coffee scenario from~\citet{DBLP:journals/ai/WinikoffSDD21} 
(see \supp{figure~\ref{fig:coffee}}).
The scenario is that the agent has the desire to \textsf{getcoffee}, and has three ways to do so: (i) get (low quality but free) coffee from a kitchen, which requires a staff card (either one's own, or a colleague's), and involves the sequence: \textsf{getStaffCard} (sub-goal), \textsf{goto(kitchen)},  \textsf{getCoffee(kitchen)} (both actions); (ii) get (decent quality free) coffee from a colleague's office, which uses pods and requires the colleague, Ann, to be in their office, and involves the sequence of actions \textsf{goto(office)}, \textsf{getPod}, \textsf{getCoffee(office)}; or (iii) get (good but expensive) coffee from a shop, which requires money, and involves the sequence of actions \textsf{goto(shop)}, \textsf{pay(shop)}, \textsf{getCoffee(shop)}.

\begin{figure*}[ht]
\begin{framed}
\begin{eqnarray}
   \E^T_{X\conts{/F}} &=&
        \{ \mathsf{D{:}}N \;|\; \ancestor_{\conts{F}}(N,X) \land \lnot isOne(N) \land \lnot isXOne(N) \land \lnot isSOne(N) \}  \label{eq:d}
      \\ & & {} \cup \{ \mathsf{B{:}}\mathit{filter}(A_i) \;|\; \orig{i \leq j} \; \cont{i = j} \}  \label{eq:b1}
      \\ & & {} \cup \{ \mathsf{B{:}}\cond(N) \;|\; N=X \lor \ancestor_{\conts{F}}(N,X) \}  \label{eq:b2}
      \\ & & {} \cup \{ \mathsf{B{:}}\lnot\cond(N_i) \;|\; \ancestor^{\conts{ca}}_{\conts{F}}(N,X) \land isOne(N) \land 
            sib(N,X,\_,N_i) \land nheld(N_i) \}  \label{eq:b3}
    \\ & & {} \cup \{ \mathsf{B{:}}\lnot\cond(N_i) \;|\;  \ancestor^{\conts{ca}}_{\conts{F}}(N,X) \land isSeqOne(N) \\
    & & \hspace*{1cm}  {} \land sib(N,X,N_x,N_i)  \land \cont{sib(N,F,N_f,\_) 
            \land  seqn(N_f) \leq } seqn(N_i)<seqn(N_x)  \} \label{eq:b4} 
      \\ & & {} \cup \{ \mathsf{V{:}}{N_i}{<}{N_x} \;|\;  \ancestor^{\conts{ca}}_{\conts{F}}(N,X) \land isOne(N) \land 
            sib(N,X,N_x,N_i) \land held(N_i) \}  \label{eq:v}
      \\ & & \mbox{Where } T = \langle A_1  \ldots A_j=X  \ldots A_n \rangle \\
      \mathit{filter}(A_i) &=& \pre(A_i) \setminus \{ c \; |  \; \seqb(N,A_i) \land isAct(N) 
      \land N \in T \land c \in \mathit{post}(N) \} \label{eq:filter} \\
sib(N,X,N_x,N_i) &=& child(N_x,N) \land (\ancestor(N_x, X) \lor N_x=X) \land child(N_i,N) \land x \neq i \label{eq:sib2} \\
\seqb(N_1,N_2) &=& \begin{array}[t]{l}
(\ancestor(N'_1,N_1) \lor N'_1=N_1) \land
(\ancestor(N'_2,N_2) \lor N'_2=N_2) \label{eq:seqb} \\  {} \land 
parent(N'_1)=parent(N'_2) \land isSeq(parent(N'_1)) 
    \land seqn(N'_1) < seqn(N'_2) \label{eq:seqb2} \end{array} \\
 \cont{\ancestor_F(N,X)} & \cont{=} & \cont{\ancestor(N,X) \land \lnot \ancestor(N,F)} \label{eq:ancf} \\
  \cont{{{\ancestor}_F^{ca}}(N,X)} & \cont{=} & \cont{\ancestor_F(N,X) \lor ca(N,X,F)} \label{eq:anccaf} \\
\cont{ca(C,A,B)} &\cont{=}& \cont{\ancestor(C,A) \land \ancestor(C,B) \land 
\lnot \exists N . \ancestor(N,A) \land \ancestor(N,B) \land \ancestor(C,N) } \label{eq:ca} \\ 
\cont{\E^T_{X/?}} &\cont{=}& \cont{\displaystyle \bigcup_{f \in \mathit{vf}(X)} \E^T_{X/f}} \label{eq:implicit} \\
\cont{\mathit{first}(N)} & \cont{=}& 
\cont{
	\begin{cases}
	\{N\}, & \text{if } N.\mathit{type} = Action \\
	\bigcup_{N_i \in \mathit{children}(N)} \mathit{first}(N_i), & \text{if } N.\mathit{type} \neq seq \\
	\mathit{first}(N_1), & \text{if } N.\mathit{type} = seq \land  child(N_1,N) \land seqn(N_1) = 1 
	\end{cases}} \label{eq:first} \\
\cont{\textit{vf}(X)} & \cont{=} & \cont{\begin{array}{l}
	\{ F \; | \; 
	ca(C,X,F) \land ( isOne(C) \lor isXOne(C) \lor isSOne(C) ) \label{eq:vf1} \\ 
	 {} \land child(N_X,C) \land (\ancestor(N_X,X) \lor N_X=X) \land X \in \mathit{first}(N_X) \label{eq:vf2} \\
	 {} \land child(N_F,C) \land (\ancestor(N_F,F) \lor N_F=F) \land F \in \mathit{first}(N_F) \;
	\}  \label{eq:vf3}  \end{array}} 
\end{eqnarray}    

\end{framed}
\caption{Definition of $\E^T_X$ (ignore green shaded parts) and of $\E^T_{X/F}$ (adding $\cont{\mbox{green shading}}$).
Auxiliary functions and predicates used: $\pre(N)$ and $post(N)$ (return pre-/post-condition of an action as a conjunction represented as a set of propositions),  $\cond(N)$  (returns the condition of node $N$),  $parent(N)$ (return the parent node of $N$ in the given goal tree),  $children(N)$ (return the children of $N$), $seqn(N)$ (returns the sequence number of $N$),  $\ancestor(A,B)$  (true when $A$ is an ancestor of $B$), $child(N_i,N)$ (true when $N_i$ is a child of $N$),  $isOne(N)$ (resp.~$isXOne$, $isSOne$, and $isSeq$) which is true when the type of node $N$ is $one$ (resp.~$sone$, $xone$, $seq$), and   $isAct(N)$ (true when $N$ is an action node). We also use the predicate $held(N)$ which is true when the condition of node $N$ held when the node was reached (or if there is no condition), and $nheld(N)$ which is true when the condition of node $N$ did not hold.
}\label{fig:explain}
\end{figure*}

\subsection{Explanation Generation}\label{sec32}

We view an explanation  as a set of time-tagged \emph{explanatory factors} $F$, where each factor is either a \emph{desire}, a \emph{belief},  or a \emph{valuing}. Formally: 
$ F ::= \mathsf{D{:}}_{t}N \;|\; \mathsf{B{:}}_{t}C \;|\; \mathsf{V{:}}_{t}N_1{<}N_2 $
where $N$ is the name of a node,  $C$ is the condition of a node or a pre-condition of an action, and $N_1{<}N_2$ indicates that $N_2$ is preferred over $N_1$, and $t$ is a time tag which we omit in the remainder of the paper.
We  define  $\E^T_X$ as being the set of explanatory factors that explain action $X$ with regard to trace (i.e.~sequence of actions) $T$ and in the context of the given goal tree. We assume (not formalised below) that $X \in T$, otherwise the response to ``why did you do $X$?'' is ``I didn't''.

The basic idea is that to explain ``why did you do $X$?'' we consider the three types of explanatory factors.
 \textbf{(1) Desires:}  What was the goal that was trying to be achieved? This can be found by considering the ancestors of $X$ in the tree. However, we filter out \textsc{Or} nodes because they provide less information than their child, for example, ``I want to get coffee'' is less informative than ``I want to get office coffee'' (Figure~\ref{fig:explain}, Line~\ref{eq:d}; ignore green shaded parts).
\textbf{(2) Beliefs:} What conditions held (or failed to hold) that were relevant to choosing to do $X$? There are three cases here. The first two cases relate to why we were \emph{able} to do $X$, whereas the third (and Valuings, below) relate to why we \emph{chose} to do $X$.
\textbf{Case 1:} Pre-conditions of $X$  and of actions that were done before $X$ (i.e.~appear before it in the trace): these must hold, otherwise $X$ could not be reached and done (Figure~\ref{fig:explain}, Line~\ref{eq:b1}). 
We filter these by removing pre-conditions of $Y$ that were achieved by another action $Z$ that necessarily occurs before $Y$ in the trace (Figure~\ref{fig:explain}, Line~\ref{eq:filter}), e.g.~getting shop coffee would not include the condition that we were at the shop (pre-condition of \textsf{getCoffee(shop)}) because it is a post-condition of \textsf{goto(shop)}.
\textbf{Case 2:}        
Conditions on nodes: let $N$ be either $X$ or an ancestor of $X$, then its condition had to hold for $X$ to be reached, so is included in the explanation (Line~\ref{eq:b2}). For example, I chose to go to the shop because I had money, so that option was available.
  \textbf{Case 3:} Excluded choices: when we choose a particular child of a $one$ node, some of the other possible choices (siblings) are excluded because their conditions did not hold. This is part of the explanation for why we chose the child we did. For example, I chose to go to the shop because I could not get office coffee because Ann was not in her office. More precisely, let $O$ be a node of type $one$ that is an ancestor of $X$, and that has children $N_1, \ldots, N_x, \ldots N_n$ where $X$ is a descendant of $N_x$ or $X=N_x$. Then for those $N_i$ ($i \neq x$) where their condition is known to not hold, this condition is also part of the explanation (Line~\ref{eq:b3}). For $sone$ this is modified (Line~\ref{eq:b4}): we only include the conditions of siblings older than $N_x$ (i.e.~$N_i$ where $i<x$). For $xone$ we do not include the conditions of any siblings: the fact that the condition of $N_x$ was True implies that the conditions of $N_i$ ($i \neq x$) must be False.
\textbf{(3) Valuings:} These also relate to choices. Where we chose a particular child of a $one$ node, then the other options (siblings) that were available but weren't selected are less preferred.
More precisely, let $O$ be a node of type $one$\footnote{For sequential and exclusive \textsc{Or} we don't have more than one child as an option, so Valuings are not part of the explanation for those nodes.} 
that is an ancestor of $X$ with children $N_1, \ldots, N_x, \ldots N_n$ where $X$ is a descendent of $N_x$ or $X=N_x$. 
Then for nodes $N_i$ ($i \neq x$)
where the condition of $N_i$ held, we chose
to pursue $N_x$ instead of $N_i$ because $N_x$ was more valued (Line~\ref{eq:v})\footnote{Adding value effects to the goal-plan tree can be done straightforwardly, following~\citet{DBLP:journals/ai/WinikoffSDD21}, and allows for more precise explanations for why $N_x$ was valued more than $N_i$.} 

Figure~\ref{fig:explain} shows the formal definition of $\E^T_X$ (for now please ignore the parts that are boxed and shaded green). In addition to a number of straightforward auxiliary functions and predicates (noted in the figure's caption), we also define two predicates that are a little more complex.
The first is $sib(N,X,N_x,N_i)$ (Line~\ref{eq:sib2}) which is true when $N_i$ is a child of $N$ that is a sibling of $X$ or of $N_x$, where $N_x$ is an ancestor of $X$.
The second is $\seqb(N_1,N_2)$ (Line~\ref{eq:seqb}) which is true when $N_1$ necessarily occurs before $N_2$ in the trace (i.e.~their common ancestor is a $seq$ node and $N_1$ occurs before $N_2$ in the sequence). 

\textbf{Example:} Returning to the coffee example, consider a situation where the agent performed the sequence of actions: 
\textsf{getOwnCard}, followed by \textsf{goto(kitchen)}, and then \textsf{getCoffee(kitchen)}.
The answer to the (non-contrastive) question  ``why did you do \textsf{getOwnCard}?''---where we assume that a colleague's card is available and that one has money, but that Ann is not in her office---is the  following set of six explanatory factors: 
\begin{equation*}
\begin{gathered}
\{ \; 
\mathsf{D}{:}\textsf{getKitchenCoffee},
\mathsf{B}{:}\textsf{ownCard}, 
\mathsf{B}{:}\textsf{staffCardAvailable},
 \\
\mathsf{B}{:}\lnot\textsf{AnnInOffice}, 
\mathsf{V}{:}\textsf{getOthersCard}{<}\textsf{getOwnCard}, \\
\mathsf{V}{:}\textsf{getShopCoffee}{<}\textsf{getKitchenCoffee} \; \} 
\end{gathered}
\end{equation*}

\section{Contrastive explanation generation}\label{sec:contrastive}

In defining contrastive explanations we begin with the observation that a contrastive query is in effect a \emph{filter}. When asked ``why did you do $X$ and not $F$?'' we are still explaining why we did $X$, but the explanation is focussed on those things that explain specifically why not $F$, in other words we filter out things that don't specifically relate to explaining why not $F$.  
We next review each of the three explanatory factors to consider how the explanation generation needs to be modified to do this filtering. We then 
consider how to handle implicit contrastive questions, which requires us to first define what is a valid foil.

\textbf{Desires:} recall that these are non-\textsc{Or} ancestor nodes of the query $X$. However, any node that is an ancestor of $X$ and also an ancestor of $F$ can be filtered out, because it does not relate to explaining why not $F$.
We define this using a modified $\ancestor$ predicate (denoted $\ancestor_F$) that is only true for ancestors of $X$ that are not also ancestors of $F$ (Figure~\ref{fig:explain}, Line~\ref{eq:ancf}). Line~\ref{eq:d} of Figure~\ref{fig:explain} is then modified to use $\ancestor_F$ instead of $\ancestor$. 
 \textbf{Beliefs:} we have three cases.
 \textbf{Case 1:}   
        Pre-conditions of $X$ and of actions that were done before $X$: the pre-conditions of $X$ remain relevant. However,  the pre-conditions of earlier actions are not relevant. Specifically, following Grice's maxims~\cite{grice1975logic}, we argue that by asking the question ``why $X$ and not $F$?'' the asker is indicating that $X$ is the earliest action which was unexpected. Hence, we assume that actions preceding $X$ are correct and not relevant. We formalise this by replacing $i \leq j$ (indicated with a dashed box in Line~\ref{eq:b1}) with $i = j$. 
 \textbf{Case 2:}       
        Conditions of ancestor nodes: Similarly to the case for Desires, we omit ancestor nodes of $X$ that are also ancestor nodes of $F$  (Line~\ref{eq:b2} uses $\ancestor_F$ instead of $\ancestor$).
 \textbf{Case 3:} Choice-related non-holding conditions: Similarly to the case for desires, we focus on ancestors that are not also ancestors of $F$, except that we do want to also consider the closest common ancestor, because this is the point where a decision was made that related to the choice between $X$ and $F$. 
    We therefore (Lines~\ref{eq:b3} \& \ref{eq:b4}) use $\ancestor^{ca}_F$ instead of $\ancestor$, where 
$ \ancestor^{ca}_F(N,X) $ is defined (Line~\ref{eq:anccaf}) as ancestors of $X$ that are not also ancestors of $F$, but including their closest common ancestor (identified using  $ca(C,A,B)$, Line~\ref{eq:ca}).
Additionally, for $sone$ we also filter out explanatory factors associated with $N_i$ that appear earlier than $N_f$: since they appear before $N_f$, they cannot relate to the difference between doing $X$ and doing $F$ (Line~\ref{eq:b4}, $seqn(N_f) \leq seqn(N_i)$). 
\textbf{Valuings:} 
Similarly to the earlier cases, we exclude nodes that are also ancestors of $F$. This is done (Line~\ref{eq:v}) by using $\ancestor^{ca}_F$ instead of $\ancestor$.
Figure~\ref{fig:explain} shows the  formal definition of the explanatory factors for ``why did you do $X$ instead of $F$?'' (given tree $G$ and trace $T$) (denoted $\E^T_{X/F}$).

\textbf{Example:} Returning to the coffee example with the same sequence of actions as in \S\ref{sec32},
the answer to the contrastive question ``why did you do \textsf{getOwnCard} instead of \textsf{getOthersCard}?''  ($\E^T_{\textsf{getOwnCard}/\textsf{getOthersCard}}$) is the following set of two explanatory factors: 
$ \{ 
\mathsf{B}{:}\textsf{ownCard}, 
\mathsf{V}{:}\textsf{getOthersCard}{<}\textsf{getOwnCard} \} $.

\textbf{Implicit foils:} As noted earlier, sometimes a contrastive question has an \emph{implicit foil}, i.e.~the foil is implied rather than explicitly stated. We handle this by identifying all possible foils, and then taking the weakest (least restrictive) filter: we only filter out an explanatory factor if it is filtered out for all possible foils. 
This corresponds to taking the  \emph{set union} of the explanations for each of the foils: if a factor is in an explanation for any foil, it is included. 
We therefore define the answer to an implicit contrastive question, denoted $\E^T_{X/?}$, as the union of $\E^T_{X/f} $ over the set of possible valid foils $\mathit{vf}(X)$ (defined below), yielding Line~\ref{eq:implicit} in Figure~\ref{fig:explain}.

\textbf{Valid foils:} We now need to consider the question of what is a valid foil and define $\mathit{vf}(X)$. 
Intuitively, a valid foil $F$ is one that was not done (i.e.~$F \not\in T$), but could have been done, replacing the fact $X$.
Unfortunately implementing this intuition would require exploring possible traces to check that there exists one where $F$ is done and replaces $X$.
We therefore instead use the condition that the fact $X$ and prospective foil $F$ have a common ancestor node $C$ that is an \textsc{Or} node, and that $X$ and $F$ are both possible first actions (Figure~\ref{fig:explain}~Line~\ref{eq:first}) of the relevant child of $C$. 
This is sufficient to ensure the desired intuitive conditions are met (see Line~\ref{eq:vf1}).

\section{Computational evaluation}\label{sec:eval}

It is clear that contrastive explanations are shorter, but not by how much.  We assess the length reduction by generating 1000 random trees of sufficient size ($\geq \theta$) and comparing the size of non-contrastive and contrastive explanations for each action node $N$ in the tree and for each of its valid foils. Since the explanation generation function $\E$ requires  a trace $T$, we also generate a trace for each $N$ (see \S\ref{sec:tracegeneration}). 
Pseudo-code summarising the process can be found in \supp{\S\ref{appx:compeval}}.

\subsection{Tree Generation}\label{sec:treegen}

We follow the ``traditional'' BDI goal-plan tree structure which alternates \textsc{And} and \textsc{Or} nodes, and only permits children of \textsc{Or} nodes to have conditions. This means that an \textsc{Or} node has non-\textsc{Or} nodes as children, so if the children of an \textsc{Or} (i.e.~$one$, $sone$ or $xone$) are not action nodes, then they must be either $all$ or $seq$ (selected randomly).  Similarly, the children of \textsc{And} nodes (i.e.~$seq$ or $all$) are either action nodes or \textsc{Or} nodes (with type selected randomly from $one$, $sone$, $xone$).
A generated tree is either an action node (with  probability $\alpha$, as long as it is at depth $d>1$), or a goal node with $n$ children ($2 \leq n \leq \epsilon$, where $\epsilon$ is a parameter of the tree generation) and a type (one of $one$, $sone$, $xone$, $seq$, $all$).  Children of a $sone$ or $seq$ node have a sequence number.  All action nodes have a pre-condition, and each node that is the child of an \textsc{Or} node has a condition.  We limit the depth of the tree to $\delta$: if the depth is reached during the tree generation process then we only generate children that are action nodes.
Python code implementing tree generation is in \supp{\S\ref{appx:treegenerate}} along with brief discussion of how our tree generation differs from \citet{DBLP:conf/emas/YaoW21}.

\begin{figure*}[t!]
\centering
\fbox{
\begin{tabular}{cc}
\includegraphics[width=0.47\linewidth]{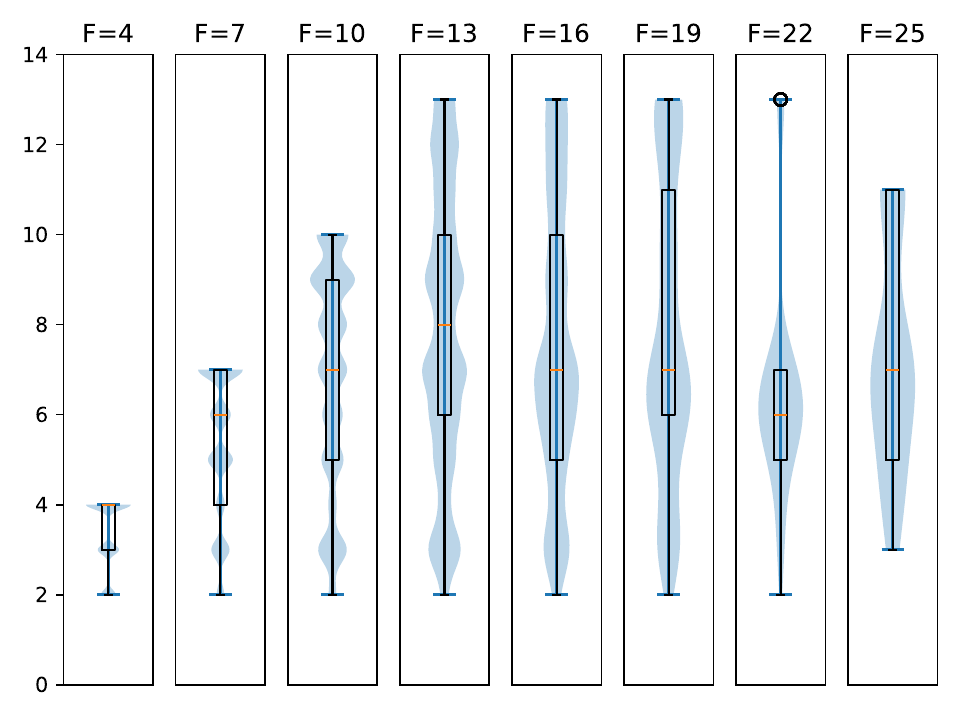} &
\includegraphics[width=0.47\linewidth]{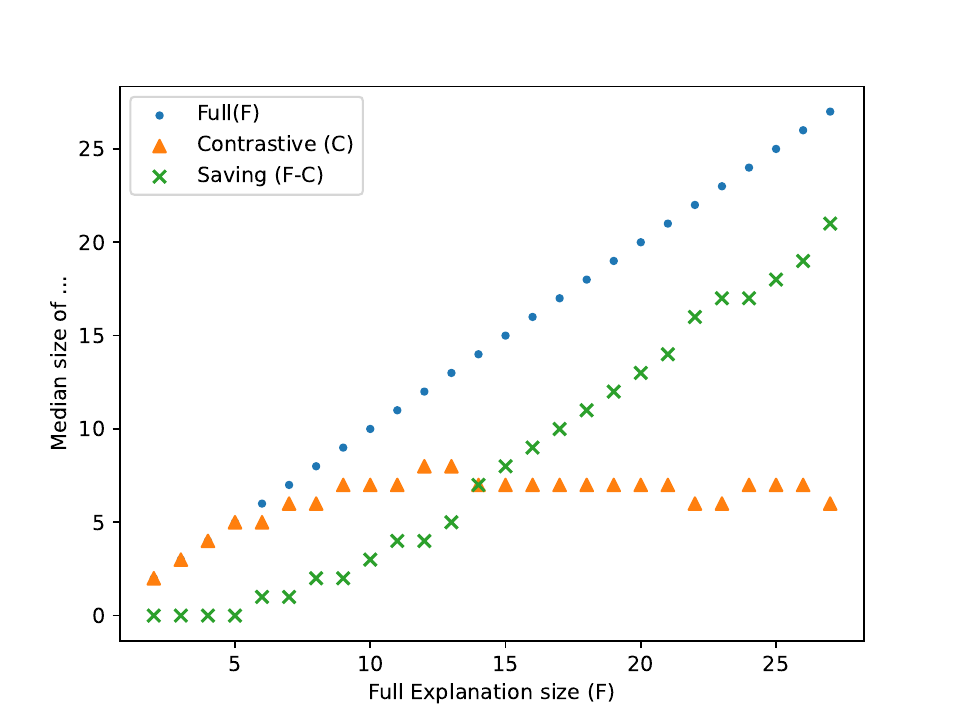} 
\end{tabular}
}
\caption{Computational Evaluation Results: plot of distribution of contrastive size against full explanation size for selected $F$ (left) and of median full size / contrastive size / saving against full explanation size (right). }\label{fig:computationalevalresultsall4}
\end{figure*}

\subsection{Trace Generation}\label{sec:tracegeneration}

We need to be able to generate a trace from a given tree $\mathcal{R}$ and a selected node $N \in \mathcal{R}$. 
In the process of generating the trace, we assign truth values to conditions in a way that is consistent with $N$ being selected. 
This assignment is needed because the explanation generation uses these ($held$, $nheld$).
We therefore define a procedure generate\_trace$(R, N)$, where $R$ is a node (initially the root of tree $\mathcal{R}$), that returns a trace and annotates nodes in $\mathcal{R}$ to indicate whether their conditions hold.

The process of generating a trace is recursive, starting with the root of the tree. 
If we reach an action node $A$ then we mark the node's condition as being True and return the trace $\langle A \rangle$. 
Otherwise, we proceed as follows, depending on the type of $R$, which we assume has children  $N_1 \ldots N_n$.

If $R$ is an \textsc{Or} node then we first select  the node $N_x$ ($1 \leq x \leq n$) such that $N_x$ is either equal to the node $N$ for which an explanation will be generated, or is an ancestor of $N$. We then mark the condition of $N_x$ as True. For $R$ of type $one$ we mark the conditions of sibling nodes $N_i$ ($i \neq x$) randomly (nodes without conditions can only be marked True).  For $R$ of type $xone$ we mark the conditions of $N_i$ as False (following the exclusive-or semantics of $xor$), and for $R$ of type $sone$ we mark nodes $N_i$ ($i < x$) as False (following the sequential-or semantics of $sor$). We then return the trace generated from $N_x$ (generate\_trace$(N_x, N)$).

If $R$ is an \textsc{And} node then we generate the traces $T_i$ recursively for each $N_i$, and then concatenate these traces sequentially (for $R$ of type $seq$) or interleave them randomly (for $R$ of type $all$). We use $\fatsemi$ to denote the function that joins two traces in sequence and  $\interleave$ to denote the function that randomly interleaves two traces.

\noindent 
\begin{mdframed}
\begin{tabbing}
X \= X \= X \= X \= \kill
\textbf{def} generate\_trace$(R,N)$: \\
\> \textbf{if} $isAct(N)$ \textbf{then}  \\ 
\> \> mark($N$,True) \\ 
\> \> \textbf{return} $\langle N \rangle$ \\
\> \textbf{if} $\mathit{isOr}(R)$ \textbf{then} \textit{// find $N_x$} \\
\> \> \textbf{for} $N_i \in \mathit{children}(R)$:  \\
\> \> \> \textbf{if} $\mathit{ancest}(N_i,N) \lor N_i=N$ \textbf{then} $N_x = N_i$ \\
\> \> mark($N_x$, True) \\
\> \> \textbf{for} $N_i \in (\mathit{children}(R) \setminus \{N_x\})$: \\
\> \> \> \textbf{if} $\mathit{isOne}(R)$ \textbf{then} markRandom($N_i$) \\
\> \> \> \textbf{if} $\mathit{isXOne}(R)$ \textbf{then} mark($N_i$,False) \\
\> \> \> \textbf{if} $\mathit{isSOne}(R) \land \mathit{seqn}(N_i){<}\mathit{seqn}(N_x)$ \textbf{then} \\ \> \> \> \> mark($N_i$, False) \\
\> \> \textbf{return} generate\_trace$(N_x,N)$ \\
\> \textbf{if} $\mathit{isAnd}(R)$ \textbf{then} \\
\> \>  \textbf{for} $i \in \{1 \ldots n\}$:  $T_i \leftarrow \text{generate\_trace}(N_i,N)$ \\
\> \> \textbf{if} $\mathit{isAll}(N)$ \textbf{then} \textbf{return} 
	$T_1 \interleave \ldots \interleave T_n$ \\
\> \> \textbf{if} $\mathit{isSeq}(N)$ \textbf{then} \textbf{return} 
	$T_1 \fatsemi \ldots \fatsemi T_n$ 
\end{tabbing}
\end{mdframed}

\subsection{Results}\label{sec:resultscomp}

We generated 1000 trees using the following generation parameter values: 
$\alpha = 0.5$ (probability of a node being an action node), 
$\delta = 5$ (maximum tree depth), 
$\epsilon = 5$ (maximum number of children), 
and $\theta = 20$ (minimum tree size). 

These parameter values are based on the goal-plan tree
of~\citet{DBLP:conf/mates/BroekensHHBJM10}.
This tree has 22 nodes (consistent with
$\theta=20$), and each non-leaf node has 2-5 children (hence $\epsilon=5$). 
The tree has 14 actions, of which 9 are at the bottom of
the tree (depth of 4), so considering nodes that are \emph{not} at the maximum
depth, for which $\alpha$ applies, we have 5 action nodes and 7 non-action
nodes (4 at depth 3 and 3 at depth 2), which is consistent with $\alpha = 0.5$.
Finally, this tree has a maximum depth of 4. However, it does not follow the
traditional BDI structure of alternating \textsc{Or} and \textsc{And}, and we
therefore increase $\delta$ to $5$, since having an \textsc{Or} node with an \textsc{Or} node as a child, say, would require an additional intervening \textsc{And} node in the traditional BDI structure.

Figure~\ref{fig:computationalevalresultsall4} shows the evaluation results.
We use $F$ to refer to the size of the Full (non-contrastive) explanation and $C$ to refer to the size of the Contrastive explanation. Since the potential saving in size ($F-C$) is limited by the size of $F$, we plot (left side of Figure) the distribution of $F-C$ for different selected values of $F$. This shows that, especially for larger $F$, the saving by using contrastive explanations can be significant. We also plot the median values of $F$, $C$ and $F-C$ for each value of $F$ (right side of Figure). This shows that as the original explanation size ($F$, x-axis) increases beyond 10, the median size of the contrastive explanation ($C$,  ``\textcolor{myorange}{$\blacktriangle$}'')  remains roughly fixed, and hence the saving ($F-C$,  ``\textcolor{darkgreen}{\textsf{x}}'') grows. 

Additional results can be found in \supp{\S\ref{appx:comp_more_results}}, including results for two example trees from the  Intention Progression Competition\footnote{\url{https://gitlab.com/intention-progression-competition/example-gpts/}}, which also show substantial length reductions ($C$ vs.~$F$).

\section{Human Subject Evaluation}\label{sec:human_subject_evaluation}

We have shown that contrastive explanations are significantly shorter. We would therefore expect that contrastive explanations would have lower cognitive load, since they are shorter, and hence be both \emph{preferred} and more \emph{effective}. To assess this we conduct a human subject evaluation, specifically answering the two questions:
(1) are contrastive explanations  \emph{preferred}?  (2) are they \emph{effective} (at supporting appropriate trust and transparency)? These are distinct questions: people may prefer explanations that are less effective (e.g.~\cite{DBLP:conf/aaai/AmitaiSA24}).

There has been a range of work on evaluating explanations~\cite{DBLP:conf/mates/BroekensHHBJM10,DBLP:conf/iat/HarbersBM10,DBLP:conf/extraamas/WinikoffS23,DBLP:journals/aamas/AbdulrahmanRB22,DBLP:conf/acii/KapteinBHN19,DBLP:conf/ro-man/KapteinBHN17,9575917,DBLP:conf/chi/GyevnarDQCBLA25,DBLP:conf/aaai/AmitaiSA24}, but none of it addresses our research questions, mostly because contrastive explanations were not considered.  
\citet{9575917} found that in the domain of autonomous vehicles participants given contrastive  explanations did not have a significant difference in understanding compared to those receiving ``Why'' explanations, but that they were significantly better at assigning accountability for a collision or near miss.

\subsection{Research Hypotheses}\label{sec:research_hypotheses}

We have 9 hypotheses. H1-H3 relate to the \emph{preference} between contrastive and full explanations, H4-H6 relate to the \emph{effectiveness} of explanations, H7-H8 concern the difference between providing and not providing explanations at all, and H9 looks at the relationship between trust in the specific system at hand and general trust in technology.
We hypothesise that, since people tend to ask contrastive questions, and these give shorter more focused answers, that these would be preferred to full explanations (\textbf{H1}), have higher \emph{perceived quality} (\textbf{H2}),  be more likely to be seen as having the right level of detail (\textbf{H3}),  yield higher trust (\textbf{H4}), higher belief in understanding of the system (\textbf{H5}), and higher confidence in the system's correct behaviour (\textbf{H6}). We also hypothesise that providing an explanation (either contrastive or full) yields higher trust  (\textbf{H7}) and more confidence in the system's correct behaviour (\textbf{H8}) than not providing an explanation. Finally, we hypothesise (\textbf{H9}) that there is a correlation between trust in technology in general and trust in each of the two systems, but that the strength of the correlation is not high.

\subsection{Methodology}\label{sec:methodology}

We recruited participants\footnote{The experiment had ethics approval from the human ethics committee of Victoria University of Wellington (approval number 2024/HE000068).} using the Prolific platform, requesting a gender-balanced sample of adults (18 years or older) who are fluent in English. 
Each participant was randomly allocated in a balanced way to one of three groups (\textbf{X}): FULL explanation, CONTrastive explanation or no explanation (``NONE''), and asked to complete a survey.

The survey (see \supp{\S\ref{appx:survey}})
had the following sections: 
\textbf{(1) Consent.}
\textbf{(2) Technology Trust (TT):} we measured the participant's general trust in technology, adopting the questions used by \citet{DBLP:conf/extraamas/WinikoffS23}, which are based on \citet{McKnight2011}, responses were on a 7 point Likert scale. 
\textbf{(3) System presentation:} Participants were presented with a high-level description of a system (see \supp{\S\ref{appx:systems}}): the first system was a robot making pancakes (based on~\citet{DBLP:conf/mates/BroekensHHBJM10}), the second was a search-and-rescue unmanned aerial vehicle (UAV).
\textbf{(4) Scenario presentation:} For each system, three scenarios for that system were presented (\S\ref{appx:systems}), each giving the current situation, indicating what the system did, and then giving no explanation or a full explanation or a contrastive explanation\footnote{The explanations were generated by software from a goal-plan tree and manually rendered in English.}. 
\textbf{(5) Explanation Quality (Q):} For each scenario,  participants who received an explanation were asked questions about the quality of the explanation, adopting the  instrument from \citet[Table 3]{DBLP:journals/fcomp/HoffmanMKL23}, 
as well as a question about understanding (\textbf{U})  following \citet{DBLP:conf/iui/BucincaLGG20}, and about the level of detail (\textbf{LD}) of the explanation being too little, about right, or too much.
 Responses (Q \& U) were on a 5 point Likert scale. 
\textbf{(6) Trust (Short):} For each scenario, all participants were  asked about their trust in the system (``short trust''). Because this question is repeated for each scenario, we only asked one question here: ``\textit{I trust the autonomous system}'' (response on a 7 point Likert scale). 
Participants were also asked about their belief in the correctness (\textbf{COR}) of the system: ``\textit{Do you believe that the system did what it should in this case?}'' (possible responses: Yes, No, Unsure).
\textbf{(7) Trust (Long):} After the third scenario for the system was presented, and associated questions answered (i.e.~(5) \& (6)), trust in the system (\textbf{T}$_{Pan}$ and \textbf{T}$_{SAR}$) was assessed, following \citet[Table 8]{DBLP:journals/fcomp/HoffmanMKL23}. Answers were on a  5 point Likert scale.
\textbf{(8) Preferred Explanation (PRE):} For each of the 6 scenarios, the participant was given a description of the situation and the system's selected course of action, and then asked whether they preferred the full explanation, the contrastive explanation, or whether they preferred the explanations equally. All participants were asked the same questions in this section. Responses were on a 5 point Likert scale (1 = ``I strongly prefer explanation 1'', 2 = ``I somewhat prefer explanation 1'', 3 = ``I prefer these explanations equally'', 4 = ``I somewhat prefer explanation 2'', 5 = ``I strongly prefer explanation 2''; explanation 1 was the full explanation, and explanation 2 was the contrastive one).
\textbf{(9) Demographic information:}  We used the same questions as Sidorenko \& Winikoff~~\cite{DBLP:conf/extraamas/WinikoffS23}, asking about age, highest level of education, gender, ethnicity, and programming experience. \label{lastsurveysec}
With the two systems, and three scenarios for each system, the sequence of survey sections that participants saw was: 12 34564564567 34564564567 89 (participants in the NONE group were not shown section 5 so instead saw: 12 34646467 34646467 89).

\subsection{Results}\label{sec:results}

We now describe our results. We begin (\S\ref{sec41}) with a summary of the responses received, and the checking and filtering that was done, as well as demographics. We then proceed to present the results, beginning with an analysis of participants' \emph{preferences} between contrastive and full explanations (\S\ref{sec42}, H1-H3), and then the \emph{effectiveness} of explanations (e.g.~resulting trust) comparing contrastive and full explanations (\S\ref{sec43}, H4-H6). We then consider the difference between having and not having explanations (\S\ref{sec44}, H7-H8), and finally we consider the relationship between trust in each of the two autonomous systems and general trust in technology (\S\ref{sec45}, H9), before finishing with discussion of the results (\S\ref{sec46}).
Table~\ref{hypothesestable} has a summary of the hypotheses, variables, tests used, and the results. 

\begin{table}[ht]
\centering
\begin{tabular}{|lllll|}
\hline
\textbf{Hyp} & \textbf{Variables} & \textbf{Test} & \textbf{Results} & \textbf{Section}  \\
\hline 
H1 & PRE & 1SW & \yes\ (partly, see text) & \multirow{3}{*}{$\Biggl\}$\S\ref{sec42}}\\ 
H2 & X$_2$-Q & M-W & \no & \\ 
H3 & X$_2$-LD & M-W & \no & \\ \hline
H4 & X$_2$-T$_{X}$ & M-W & \yes & \multirow{3}{*}{$\Biggl\}$\S\ref{sec43}} \\ 
H5 & X$_2$-U & M-W & \yes\ (scenario 2 only) & \\ 
H6 & X$_2$-COR & M-W & \yes\ (scenarios 1\&2 only) & \\ \hline
H7 & X-T$_{X}$ & K-W & \no & \multirow{2}{*}{$\Bigl\}$\S\ref{sec44}} \\ 
H8 & X-COR & K-W & \no\ (see text) & \\ \hline 
H9 & TT-T$_{X}$ & SRC & \yes & \S\ref{sec45} \\ \hline 
\end{tabular}
\caption{Summary of Hypotheses, Variables, Tests, and Results.  \textbf{Key for variables:}
X = explanation (full, contrastive, or none), X$_2$ (only full and contrastive),
T$_x$ = trust in system $x \in \{ Pan, SAR \}$, 
PRE = preference for explanation type, 
Q = quality of explanation, 
U = belief in understanding of system, 
COR = belief the system did the right thing, 
LD = level of detail of explanation (too little, about right, too much),
TT = technology trust.
\textbf{Key for tests:} 
M-W = Mann-Whitney, K-W = Kruskal-Wallis, 1SW = one-sample Wilcoxon, SRC = Spearman's rank correlation.}\label{hypothesestable}
\end{table}

\subsubsection{Responses and Filtering}\label{sec41}

We received 161 responses. These were filtered by removing low-quality responses using two mechanisms. 
Firstly, the survey included two attention check questions, which asked for agreement with a statement that was clearly false. 
Any participant who failed both attention checks was blocked from completing the survey. Participants who failed exactly one of the questions were able to complete (as per Prolific's policy on attention check questions), but their responses were not used in the analysis. Out of the 161 responses, 12  failed both attention check questions, and 18 failed exactly one of the questions. This left 131 responses.
Secondly, the survey included for each scenario a short trust question, and a longer set of questions measuring trust for each system. A number of participants had inconsistent responses for these questions. We calculated a difference score by normalising each trust measurement
to be out of 10 and then considered the difference to be too big if it was 2 or more out of 10. This check was done separately for each of the systems (pancake and search-and-rescue) and resulted in the exclusion of a further 27 participants, leaving 104 responses for analysis.
We also manually checked the responses of participants who had completed the survey unusually quickly, but these all appeared reasonable. 

For the three variables that were measured by multiple questions we calculated Cronbach's $\alpha$ to check that they were internally consistent. All were high enough ($\alpha \geq 0.8$).

The demographic profile of the 104 responses is as follows. 
Gender: 50 Male, 54 Female.
Age:  23 participants were aged 18-24, 49 were aged 25-34, 15 (35-44), 10 (45-54), 5 (55-64), 2 (65-74),  0 (75+).
Education:  22 (completed high school), 56 (completed undergrad degree),  23 (Masters), 2 (PhD), 1 (declined to answer).
Ethnicity: 40 (African), 36 (European), 8 (North American), 7 (South American),  5 (Asian), 
 3 (Other), 2 (Australian), 2 (declined), 1 (New Zealander). 
Programming experience: 38 (None), 28 (hobby), 12 (studied at high school), 12 (currently studying degree), 10 (completed degree), 4 (other).
An analysis of the variables vs.~demographic factors found a few differences\footnote{Males had a higher T$_{SAR}$ ($p{=}0.01057$, median $3.5$ vs.~$3.0$ for females), a lower U for scenario 2 ($p{=}0.00455$, mean $3.545$ vs.~$4.306$), and a lower Q for scenario 2 ($p{=}0.04198$, median $3.43$ vs.~$3.79$).  
African participants had a higher TT than Europeans ($p{=}0.01559249$,  median $5.86$ vs.~$4.91$).}, but no age-related differences were found.

\subsubsection{Explanation Type Preferences (H1-H3)}\label{sec42}

Hypothesis 1 (contrastive explanations are preferred to full) is 
\textbf{partially confirmed}: for some scenarios there was a statistically significant preference for contrastive explanations, but for others a preference for full explanations. Specifically, a one sample Wilcoxon signed rank test of the preference variable (PREF) for each scenario (looking for a difference to not having a preference, i.e.~value of 3) found statistically significant differences for scenarios 1-4
(respectively with $p=0.001596$, $p=0.0001423$, $p=0.003559$, $p=0.000008994$) and no difference for scenarios 5 and 6 ($p=0.2914$ and $p=0.8142$). 
For scenarios 1 and 2  there was a preference for contrastive explanations (median=4).
However, for scenarios 3 and 4 there was a preference for full explanations (median=2).
Scenario 5 had a (non-statistically significant) preference for contrastive explanations, and 
scenario 6 had a bimodal distribution
(See \supp{Figure~\ref{fig:h1}} for the  distribution of responses for scenarios 3, 5 and 6).
%

Hypothesis 2 (contrastive explanations have higher perceived quality) is \textbf{not confirmed}. A Mann-Whitney test  comparing the quality of explanation scores (dependent variable Q) against the explanation type (X$_2$)
did not find a statistically significant difference for any of the scenarios ($p$ values for the six scenarios were respectively: $0.1525$, $0.2246$, $1.0$,  $0.7868$, $0.805$ and $0.6825$).

Hypothesis 3 (contrastive explanations are more likely to be considered to have a level of detail (LD) that is ``about right'') is \textbf{not confirmed}. A Mann-Whitney test comparing the assessment of the level of detail (dependent variable LD) against the explanation type (X$_2$) did not find any statistically significant differences ($p$ for the scenarios respectively $0.83$, $0.45$, $0.10$, $0.95$, $0.53$, $0.47$).

\subsubsection{Effects of Explanation Type (H4-H6)}\label{sec43}

Hypothesis 4 (H4) is that contrastive explanations yield a higher level of trust than full explanations. 
This hypothesis is \textbf{confirmed}: a Mann-Whitney test comparing the dependent variable of trust in the given system (T$_{Pan}$ and T$_{SAR}$) against the independent variable of explanation type (X$_2$) shows a statistically significant difference for both systems with $p=0.008575$ for the Pancake system and $p=0.04205$ for the search-and-rescue system. On the five-point Likert scale where higher means more trusted, the median score for contrastive explanations was 3.5 for the Pancake system and 3.67 for search-and-rescue, and for full explanations was 3.0 for Pancake and 3.17 for search-and-rescue.

Hypothesis 5 (H5) is similar to H4 but with respect to the (single) question about the participants' perceived understanding of the system (U). This hypothesis was \textbf{confirmed}, but only for Scenario 2.
A Mann-Whitney test  comparing the understanding scores (dependent variable U) against the explanation type (X$_2$) found a statistically significant difference for scenario 2 only 
($p$ values for the six scenarios were respectively: $0.1193$, $0.04079$, $0.2715$, $0.3987$, $0.3663$, and $0.7789$).

Hypothesis 6 (H6) is that contrastive explanations yield more confidence in the system's correct behaviour (COR) than full explanations. This hypothesis is \textbf{confirmed}, but only for two of the scenarios. A Mann-Whitney test comparing confidence in the system's correct behaviour (dependent variable COR) against the explanation type (X$_2$) found $p=0.005351$ for scenario 1 and $p=0.04028$ for scenario 2 ($p$ values for the other scenarios: $0.06866$, $0.07875$, $0.07907$, and $0.9511$). For scenario 1 the mean responses for the contrastive and full explanation groups were respectively $2.852941$ and $2.342857$. For scenario 2 these were $2.852941$ and $2.514286$.

\subsubsection{Effects of not having Explanations (H7-H8)}\label{sec44}

Hypothesis 7 (H7) is that both types of explanation yield higher trust than no explanation. This hypothesis is \textbf{not confirmed}: a Kruskal-Wallis test with pairwise Dunn test using the Holm correction method for multiple tests found 
no significant difference  in trust (dependent variable T$_x$) between either contrastive or full compared with not providing any explanation ($X$). Interestingly, the median trust when no explanation was provided was actually \emph{higher} than the trust when a full explanation was provided (3.33 compared with 3.17 for the Pancake system, and 3.5 compared with 3 for the search-and-rescue system), but this difference was not statistically significant.

Hypothesis 8 (H8) is that having an explanation yields more confidence in the system's correct behaviour than not having an explanation. This hypothesis is \textbf{not confirmed}. In fact, there was a statistically significant difference in the \emph{opposite} direction: 
a Kruskal-Wallis test with pairwise Dunn test using the Holm correction method for multiple tests found a significant difference between the confidence in the system's correctness (dependent variable COR) and the explanation type (X) for two of the scenarios ($p=0.01046$ and $0.0124$ respectively for scenarios 3 and 4) comparing full and no explanation, with the group that had been given full explanations having  \emph{lower} responses for the system's correctness. In other words, participants who were not given an explanation at all were more positive about the system's correctness than participants who had been given full explanations (no statistically significant difference was found for contrastive vs.~no explanation).

\subsubsection{Relationship between Trust in Technology and Trust in each system (H9)}\label{sec45}

Hypothesis 9 is that there is a significant but medium-strength correlation between trust in a specific system (pancake robot or search-and-rescue, T$_x$) and trust in technology more generally (TT). This hypothesis is \textbf{confirmed}. For both specific systems, the Spearman's rank coefficient showed a statistically significant correlation ($p=0.0000000153$ for the pancake robot and $p=0.000065$ for search-and-rescue). For the pancake robot $\rho=0.52$ is interpreted as a  moderate strength relationship. For search-and-rescue $\rho=0.38$ would be interpreted as either a moderate or weak strength relationship, depending on the boundaries used\footnote{Sources vary in their definition of ``moderate strength'', 
e.g.~$0.4-0.7$ (\url{https://tinyurl.com/3vct2nuj}), 
or $0.3-0.7$ (\url{https://tinyurl.com/mr2jshkx}) or 
$0.38-0.68$ (\url{https://tinyurl.com/2uc7wxcz}). 
}.  
Graphs showing trust in each system against trust in technology in general can be found in \supp{Figure~\ref{fig:h9both}}.
Our result is consistent with the finding of \citet{DBLP:conf/extraamas/WinikoffS23} that trust in technology in general influences, but does not determine, trust in a particular system.

\subsubsection{Discussion}\label{sec46}

There are a number of interesting points to draw out of these results. 
Firstly, we saw a difference between preference and effectiveness: contrastive explanations were not consistently preferred to full explanations, but they did give higher trust in the (trustworthy) system. 
The evaluation by \citet{DBLP:conf/aaai/AmitaiSA24} also found this same outcome (albeit in a different setting). 
Secondly, we found  that providing full explanations \emph{reduced} trust in the system compared to not having any explanations. 
\citet{DBLP:conf/acii/KapteinBHN19} similarly found that providing explanations  resulted in participants being \emph{less likely} to follow their system's recommendation. They speculated that  providing information that participants already knew might reduce their adoption of the system's recommendations. 
We speculate that providing an explanation that is either too long or overly complex may result in a decrease in trust and confidence.

We also, similarly to prior evaluations~\cite{DBLP:conf/mates/BroekensHHBJM10,DBLP:conf/iat/HarbersBM10,DBLP:conf/ro-man/KapteinBHN17,9575917}, found that results were sometimes scenario-dependent. 
One particular issue that may explain the scenario-dependent results for preferences is that for contrastive explanations the explanation is given with respect to a foil $F$. However, if the participant did not consider $F$ to be a likely course of action, then the contrastive explanation might not match their expectations, and therefore not be preferred. This appears to be a plausible explanation for scenario 3 where the contrastive explanation used flipping the pancake by throwing it as foil, rather than using the spatula, and therefore the contrastive explanation filtered out the explanatory factor that the pancake was ready to be flipped.

\section{Conclusion}\label{sec:discuss}

We extended the prior work of \citet{DBLP:journals/ai/WinikoffSDD21} with contrastive explanations, which are well motivated in the literature~\cite{DBLP:journals/ai/Miller19,DBLP:journals/jair/KrarupKMLC021}. Our computational evaluation showed that contrastive explanations were significantly shorter. In particular, as the size of the  (full) explanation grows the median size of the contrastive explanation does not grow, making it scalable. Our human subject evaluation found some (scenario-dependent) evidence of preference for contrastive explanations, and stronger evidence for the \emph{effectiveness} of contrastive explanations (higher trust, better (self-assessed) understanding, and confidence in the system's correctness). We also found  that participants who had not been given explanations had a \emph{higher} level of confidence in the correct behaviour of the system for some scenarios than participants given full explanations.

\textbf{Practical Implications:} 
Firstly, when using contrastive explanations it is important to ensure that the foil matches the user's expectation. This can be done by having the user specify the foil explicitly (e.g.~``why did you do $X$ instead of $F$?''). 
Secondly, providing explanations is not risk-free. Poor quality or too-long explanations may actually reduce trust in the system. Therefore caution needs to be taken in deploying explanation facilities.
Finally, human behaviour is complex. To avoid counter-intuitive results it is important to guide development and deployment with (carefully designed) user evaluations, and to involve representative participants in an iterative development process. 

Future work includes further evaluation, with different scenarios and different systems, since our evaluation only used two simple scenarios.
Another direction for future work is making explanations interactive, which can be done following a dialogue model~\cite{DBLP:journals/aamas/DennisO22}, or as a graphical user interface.
Another direction is to extend with additional  \emph{hypothetical} question types \cite{DBLP:conf/chi/LiaoGM20} such as: 
``what-if?'' (what would happen if the situation was changed in a certain way), 
``how to be?'' (what would need to change to obtain a certain behaviour), and 
``how to still be?'' (what changes would leave the behaviour unchanged). 
Finally, this work could perhaps be made applicable to non-BDI agents by using policy graphs to model the observed behaviour of agents, and ascribing beliefs and intentions to these agents. \citet{DBLP:journals/corr/abs-2409-19038} have done this for ``why?'' questions. 



\begin{acks}
Michael Winikoff would like to thank the University of Ljubljana for hosting him on sabbatical while this paper was written.
\end{acks}


\clearpage 

\balance 

\bibliographystyle{ACM-Reference-Format} 
\bibliography{bib}

\clearpage 

\appendix 

\onecolumn

\setlength{\parindent}{0pt}
\newcommand{\somespace}{\\[-3mm]}

\section{Supplementary Material}

\begin{figure*}[h!]
\centering
\fbox{\includegraphics[width=\linewidth]{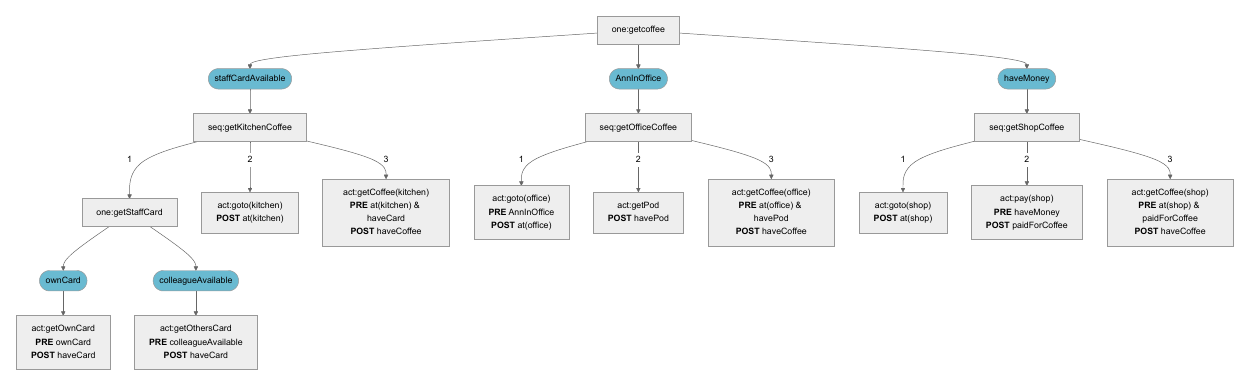}}
\caption{Coffee Example Goal-Plan Tree (based on~\cite[Figure 2]{DBLP:journals/ai/WinikoffSDD21}). 
Notation:
Nodes are rectangles that include the type and name of the goal. In the case of an action, the pre and post conditions are included in the node.
Where a node is the child of a $seq$ or $sor$ node, the sequence number is on the edge outgoing from the parent.
When a node has a condition, the condition is a ``stadium''  shaped node that sits between the goal and its parent. 
}\label{fig:coffee}
\end{figure*}

\subsection{Computational evaluation process pseudo-code}\label{appx:compeval}

\begin{mdframed}
\begin{tabbing}
X \= X \= X \= X \= X \= \kill
\textbf{repeat} 1000 times \\
\> \textbf{repeat} \\
\> \>  $\mathcal{R} \leftarrow \mbox{gen\_tree}(\text{None},\text{"or"}, 1)$ \textit{(see \S\ref{sec:treegen})}\\
\> \textbf{until} $\mathit{size}(\mathcal{R}) \geq \theta$ \\
\>  \textbf{for each} node $N \in \mathcal{R}$: \\ 
\> \> \textbf{if} $isAct(N)$ \textbf{then} \\ 
\> \> \>      $T \leftarrow \mbox{generate\_trace}(\mathit{root}(\mathcal{R}),N)$  \textit{(see \S\ref{sec:tracegeneration})} \\
\> \>  \>  \textbf{for each} $F \in \mathit{vf}(N)$: \\
\> \> \> \>    	record $size(\E^T_N)$ and $size(\E^T_{N/F})$  
\end{tabbing}
\end{mdframed}

\subsection{Python code for tree generation}\label{appx:treegenerate}

\begin{lstlisting}[language=Python,frame=single]
from anytree import Node
import random
i = 0		# global variable to generate unique names
# gen_tree(parent, type ("and", "or"), depth)
def gen_tree(par,typ,dep):
	global i
	i = i+1
	if (dep>1 and random.random() < alpha) or dep >= delta: 
		node = Node("N"+str(i), type="act", pre=["P"+str(i)],  condition=["C"+str(i)])
		if par is not None: node.parent = par
		return node
	if typ=='or': n_type = random.choice(["one","sone","xone"])
	elif typ=='and': n_type = random.choice(["all","seq"])
	node = Node("N"+str(i), type=n_type, condition = ["C"+str(i)])
	if par is not None: node.parent = par
	num_children = random.choice(list(range(2,epsilon+1)))
	if typ=="or": kids_type="and"
	else: kids_type="or"
	for n in range(1,num_children+1):
		child = gen_tree(node, kids_type,dep+1)
		if n_type in ["sone","seq"]: child.sequence = n
	return node
\end{lstlisting}

Our tree generation differs from earlier work\footnote{\url{https://github.com/yvy714/GenGPT}}~\cite{DBLP:conf/emas/YaoW21} in that 
(1) we use more node types, not just $one$ and $seq$;
(2) we allow children of a goal to be actions;
(3) each plan has $N$ children $2 \leq N \leq \epsilon$, which can be any mixture of sub-goals and actions, whereas each plan in~\citet{DBLP:conf/emas/YaoW21} has a fixed number of actions and (unless it is a leaf) a fixed number of goals; 
(4) their generation pays careful attention to generating conditions that allow the tree to be executed, which we don't need for our evaluation: for the purposes of counting, we just need to generate an arbitrary condition; and
(5) the number of children of a node varies in our generation. 
Of these differences, the last is the most significant: their trees are very regular, and, in particular, a contrastive explanation for a given tree is always the same length (an explanatory factor for each sibling of the closest $one$ node,  a single desire, and two conditions (the action, and the condition on the parent node)).

\subsection{Additional computational evaluation results}\label{appx:comp_more_results}

In addition to generating 1000 trees, our evaluation also used two example trees from the Intention Progression Competition\footnote{\url{https://gitlab.com/intention-progression-competition/example-gpts/}} (IPC): an elevator controller (``miconic'') and a logistics application. 

\begin{figure*}[h!]
\centering
\fbox{
\begin{tabular}{cc}
\includegraphics[width=0.46\linewidth]{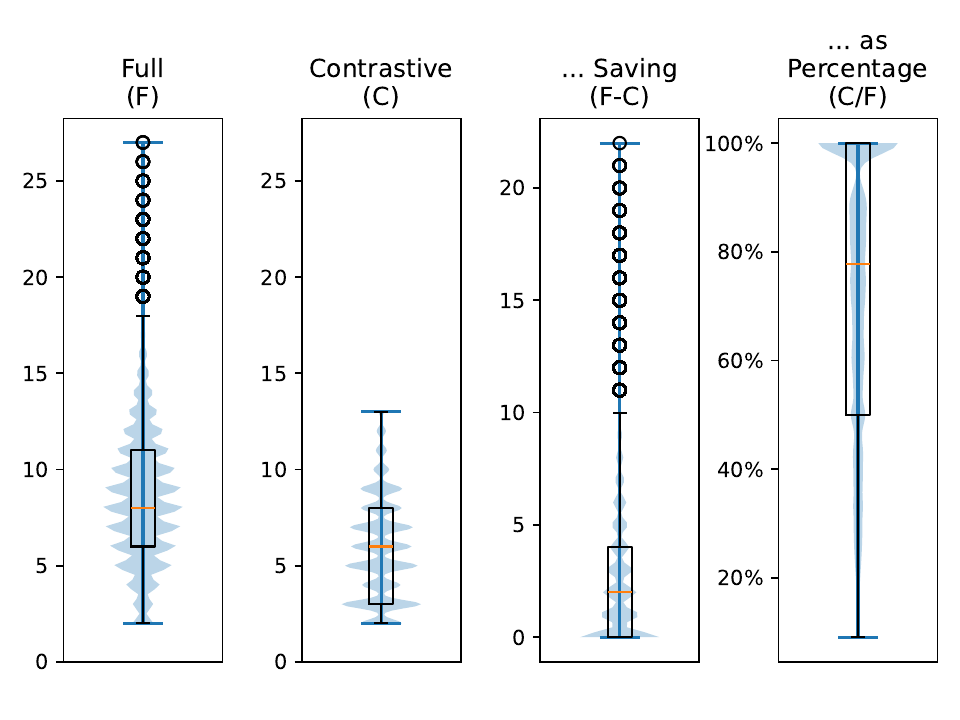} &
\includegraphics[width=0.46\linewidth]{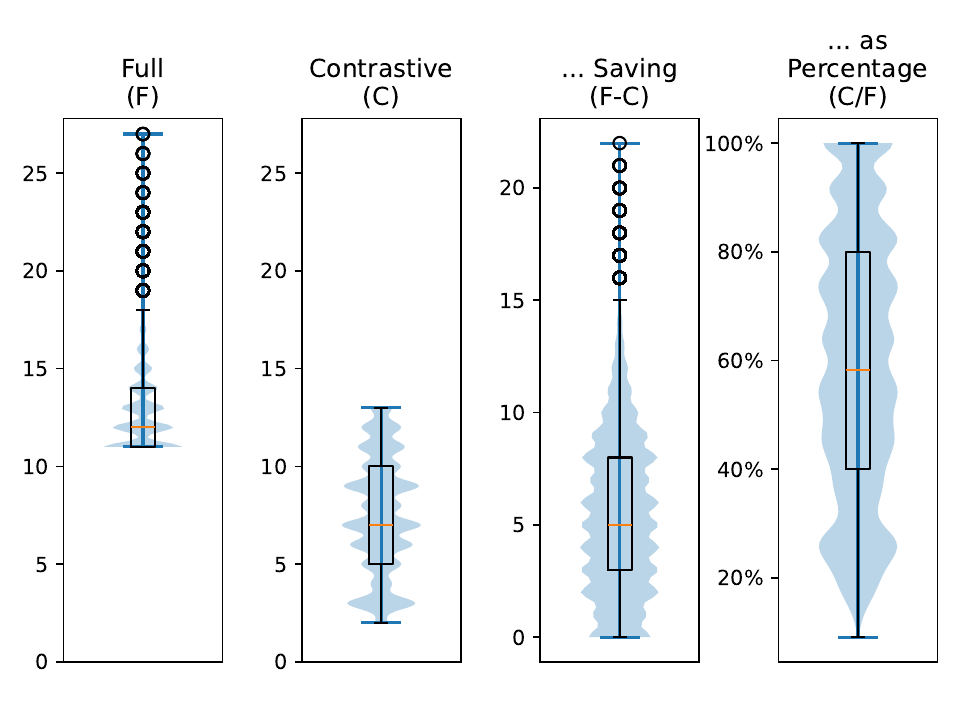}  \\
\includegraphics[width=0.46\linewidth]{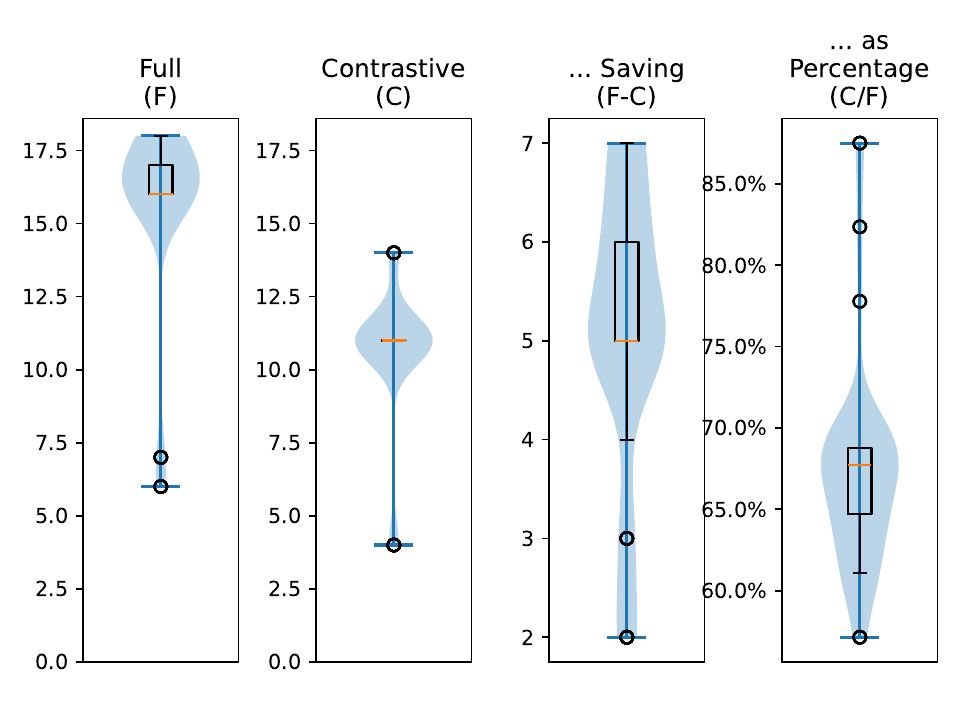} &
\includegraphics[width=0.46\linewidth]{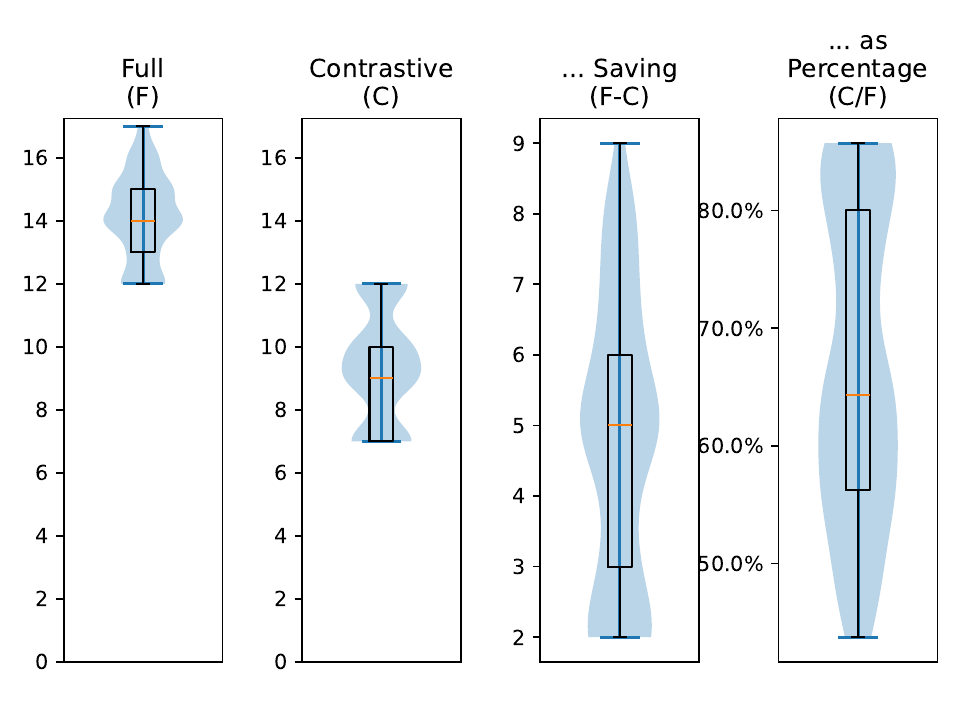} 
\end{tabular}
}
\caption{Computational Evaluation Results:  Generated Trees (top: left is all data, right $F>10$), miconic (bottom left), logistics (bottom right)  
}\label{fig:computationalevalresultsall}
\end{figure*}

The graphs (Figure~\ref{fig:computationalevalresultsall}) show the distributions (violin plot and boxplot superimposed) of both $F$ and $C$, as well as the absolute reduction in explanation size ($F-C$), and the proportional reduction in explanation size ($C/F$) obtained by using contrastive explanations. For example, for the coffee example $F=6$ and $C=2$ and hence $F-C = 4$ and $C/F = \sim 0.33$.

The  top left and bottom graphs in Figure~\ref{fig:computationalevalresultsall}  clearly show that contrastive explanations are shorter, with the difference between full and contrastive explanations varying. 
Of course, the reduction in length obtained by using a contrastive explanation is limited by the length of the full explanation to begin with. The top right graph in Figure~\ref{fig:computationalevalresultsall} shows  the results when we filter to only look at longer original explanations ($F>10)$.
This shows quite clearly that for longer explanations, the benefit (length reduction) associated with using contrastive explanations is significant: the saving ($F-C$ and $C/F$) shows clearly a significant reduction for these explanations.

\subsection{Autonomous Systems \& Scenarios}\label{appx:systems}

\textbf{System 1 description:}
The first few scenarios concern a robot making pancakes. The process for doing so involves a number of steps such as obtaining ingredients, placing them into a bowl to make the batter, heating the frying pan, and cooking the pancakes. \somespace

\textbf{System 2 description:}
The following scenarios concern unmanned aerial vehicles (UAVs) assisting in a search-and-rescue situation. The UAVs are able to search an area for people, guide (human) first responders, and deliver supplies. \somespace



\textbf{Scenario 1}

\textbf{Situation:} The robot is just starting the process of making pancakes. All the ingredients are in the house.    

\textbf{Action taken:} The robot's first action was to take the ingredients from the cupboard.

\textbf{Full Explanation:}
``I did this because: I have not collected the ingredients, I have not collected everything I need, I want to collect everything I need, I want to make pancakes, and all the ingredients are at home.''

\textbf{Contrastive Explanation:}
``I did this because all the ingredients are at home.'' \somespace

\textbf{Scenario 2}

\textbf{Situation:} The robot has collected all the ingredients and tools and has put the ingredients into a bowl. They do not have a mixer.    

\textbf{Action taken:} The robot's next action was to mix the ingredients by hand.

\textbf{Full Explanation:}
``I did this because: I want to make pancake mix, I want to make pancakes, and I do not have a mixer.''

\textbf{Contrastive Explanation:}
``I did this because I do not have a mixer.'' \somespace

\textbf{Scenario 3}

\textbf{Situation:} The robot has prepared the pancakes mixture and is currently cooking the pancake. The first side of the pancake is ready and the robot has a spatula.  

\textbf{Action taken:} The robot's next action was to flip the pancake with the spatula.

\textbf{Full Explanation:}
``I did this because: the first side of the pancake ready, I want to cook pancakes, I want to make pancakes, and I have a spatula.''

\textbf{Contrastive Explanation:}
``I did this because I have a spatula.'' \somespace

\textbf{Scenario 4}

\textbf{Situation:} The UAV accepted a request to guide a responder, and its battery has low charge. 

\textbf{Action taken:} The UAV moved towards a charging station.

\textbf{Full Explanation:}
``I did this because: my battery is low, I have a goal to support search-and-rescue operations, I am not at the charging station, and I have not received any new requests.''

\textbf{Contrastive Explanation:}
``I did this instead of charging the battery because I was not at the charging station.'' \somespace

\textbf{Scenario 5}

\textbf{Situation:}  The UAV accepted a request to deliver supplies to a first responder. It has collected the supplies, and its battery has high charge.   

\textbf{Action taken:} The UAV gave the supplies to the first responder.

\textbf{Full Explanation:}
``I did this because I am at the first responder's location and am holding the supplies, I have a task to guide the first responder, I have a goal to support search-and-rescue operations, the supplies have not yet been delivered, my battery is not low, and I have not received any new requests.''

\textbf{Contrastive Explanation:}
``I did this instead of going towards the depot because I am at the first responder's location and am holding the supplies, and the supplies have not yet been delivered.''  \somespace

\textbf{Scenario 6}

\textbf{Situation:} The UAV has accepted a request to deliver supplies to a first responder, it has not collected the supplies from the depot, and its battery has high charge.

\textbf{Action taken:} The UAV moves towards the depot.

\textbf{Full Explanation:}
``I did this because: I have a task to guide the first responder, I have a goal to support search-and-rescue operations, the supplies have not yet been delivered, I am not holding the supplies, I am not at the depot, I have not received new requests, and my battery charge is not low.''

\textbf{Contrastive Explanation:}
``I did this because: I am not holding the supplies, and I am not at the depot.''

\newpage 

\subsection{Human subject evaluation results}\label{appx:human}

\begin{figure*}[h!]
\begin{framed}
\begin{tabular}{ccc}
\includegraphics[width=0.3\linewidth]{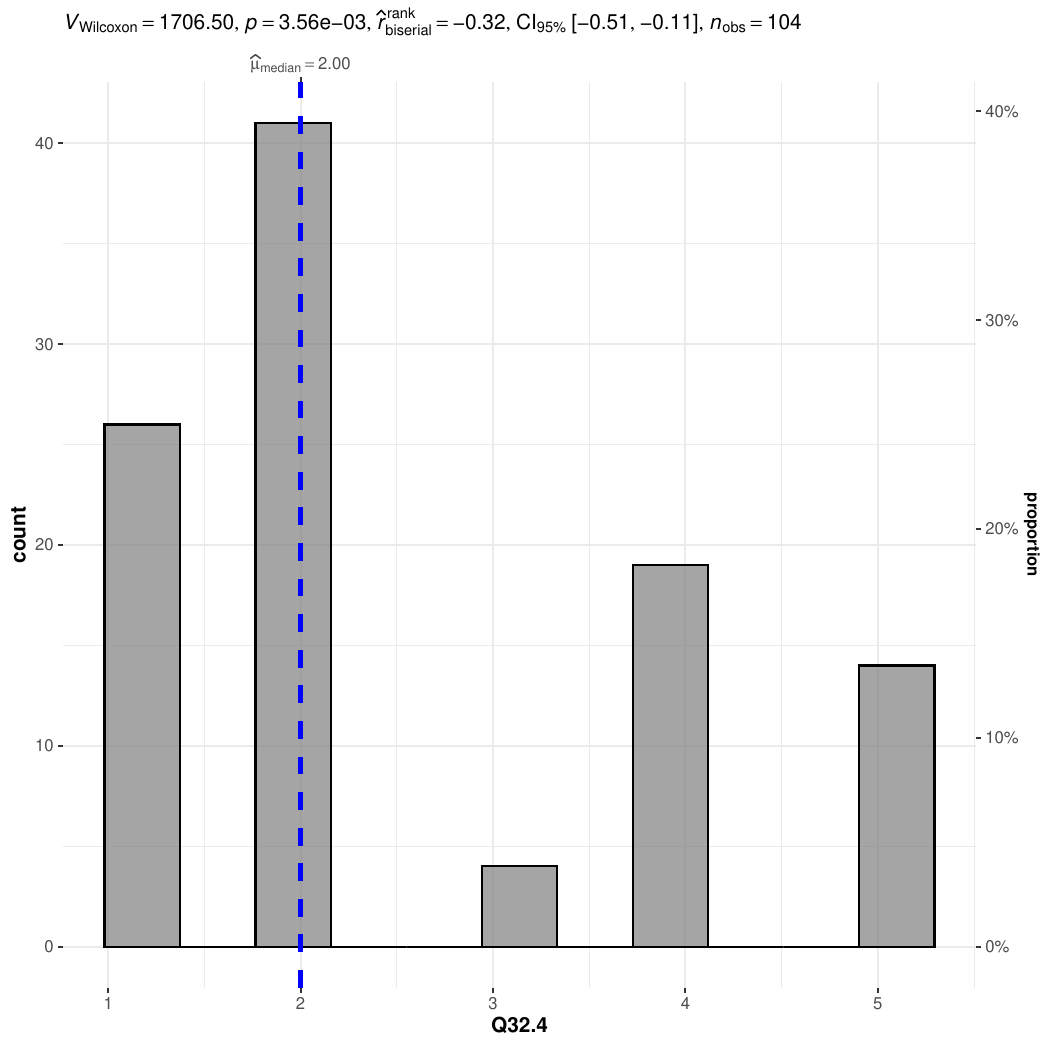} &
\includegraphics[width=0.3\linewidth]{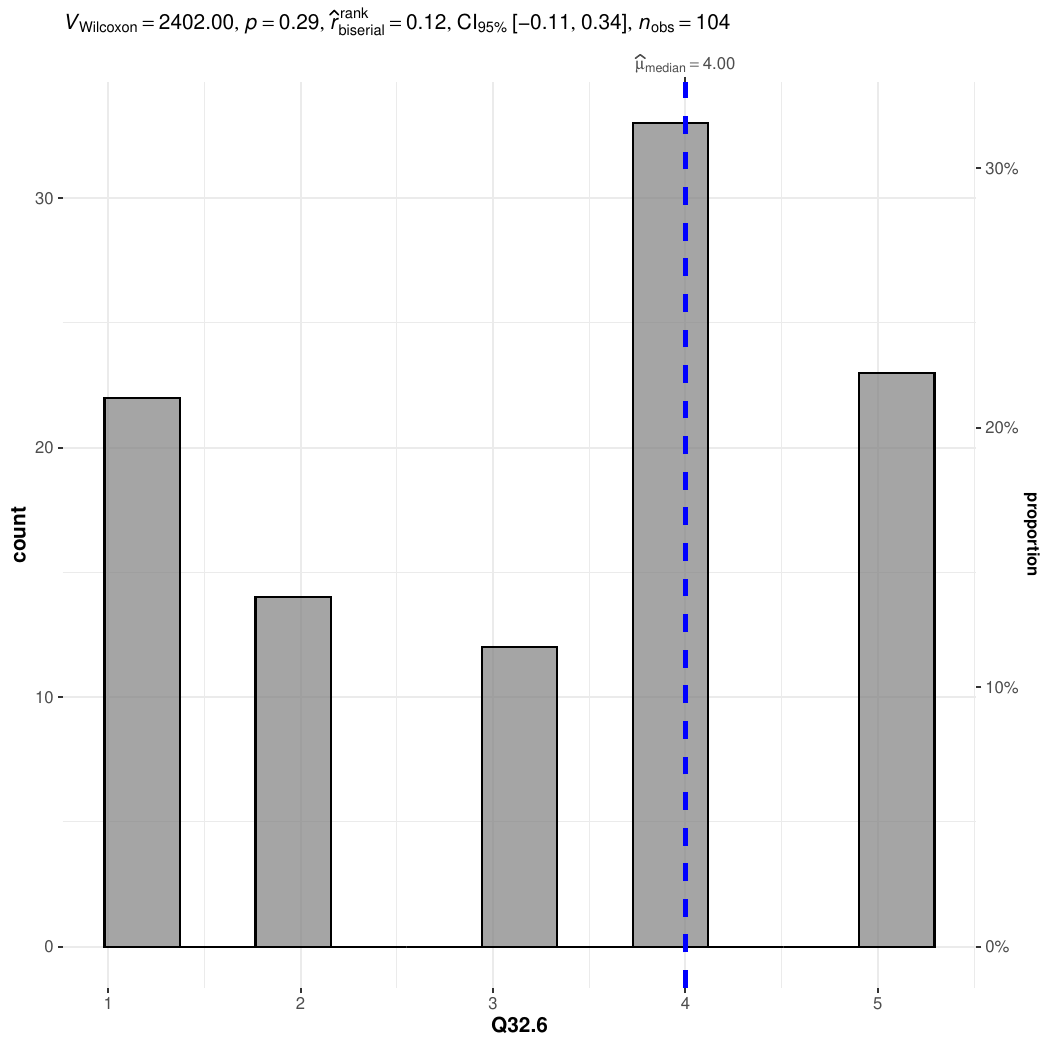} & 
\includegraphics[width=0.3\linewidth]{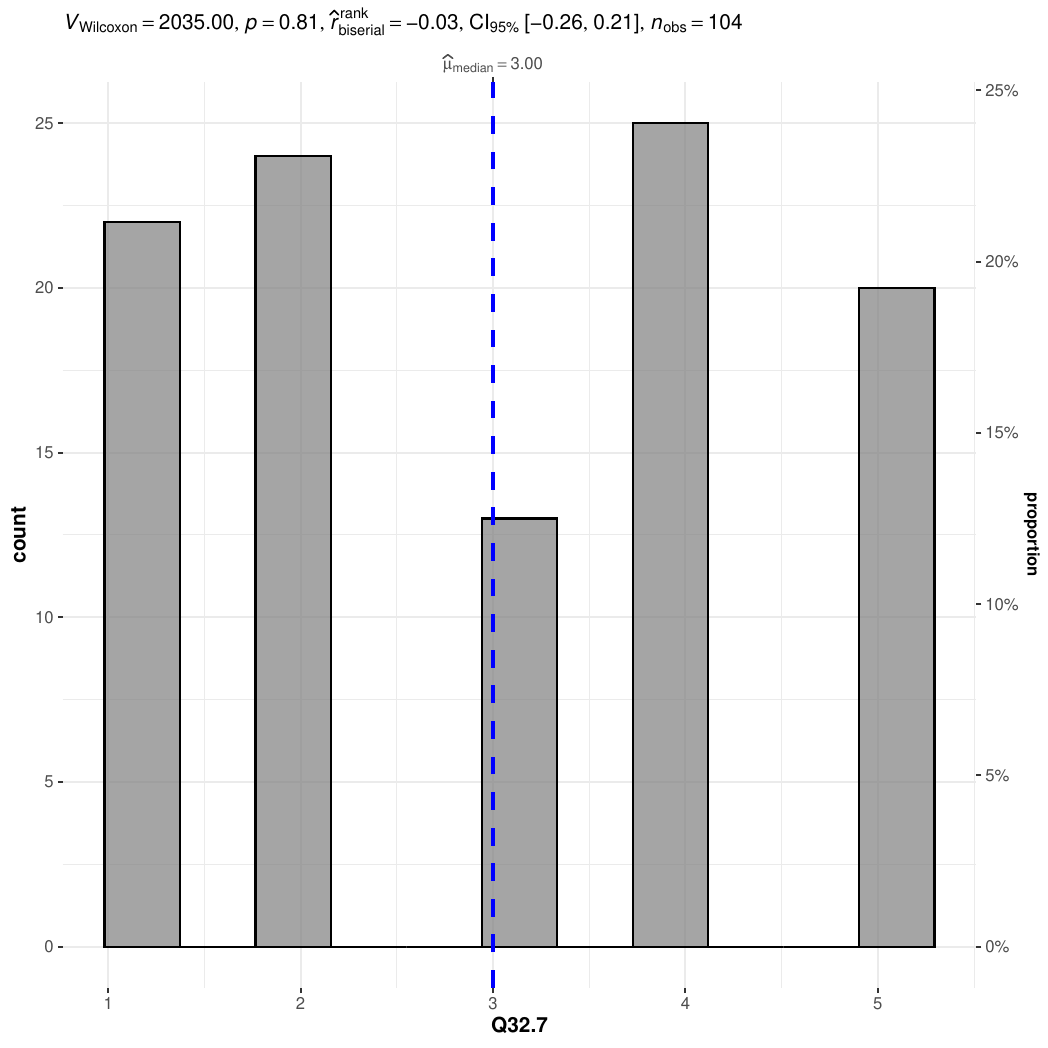}
\end{tabular}
\end{framed}
\caption{Preferred explanations for scenarios 3 (left), 5 (middle) and 6 (right); responses: 1 = strongly prefer full; 5 = strongly prefer contrastive.}
\label{fig:h1}
\end{figure*}

\begin{figure*}[h!]
\centering
\begin{framed}
\includegraphics[width=0.48\linewidth]{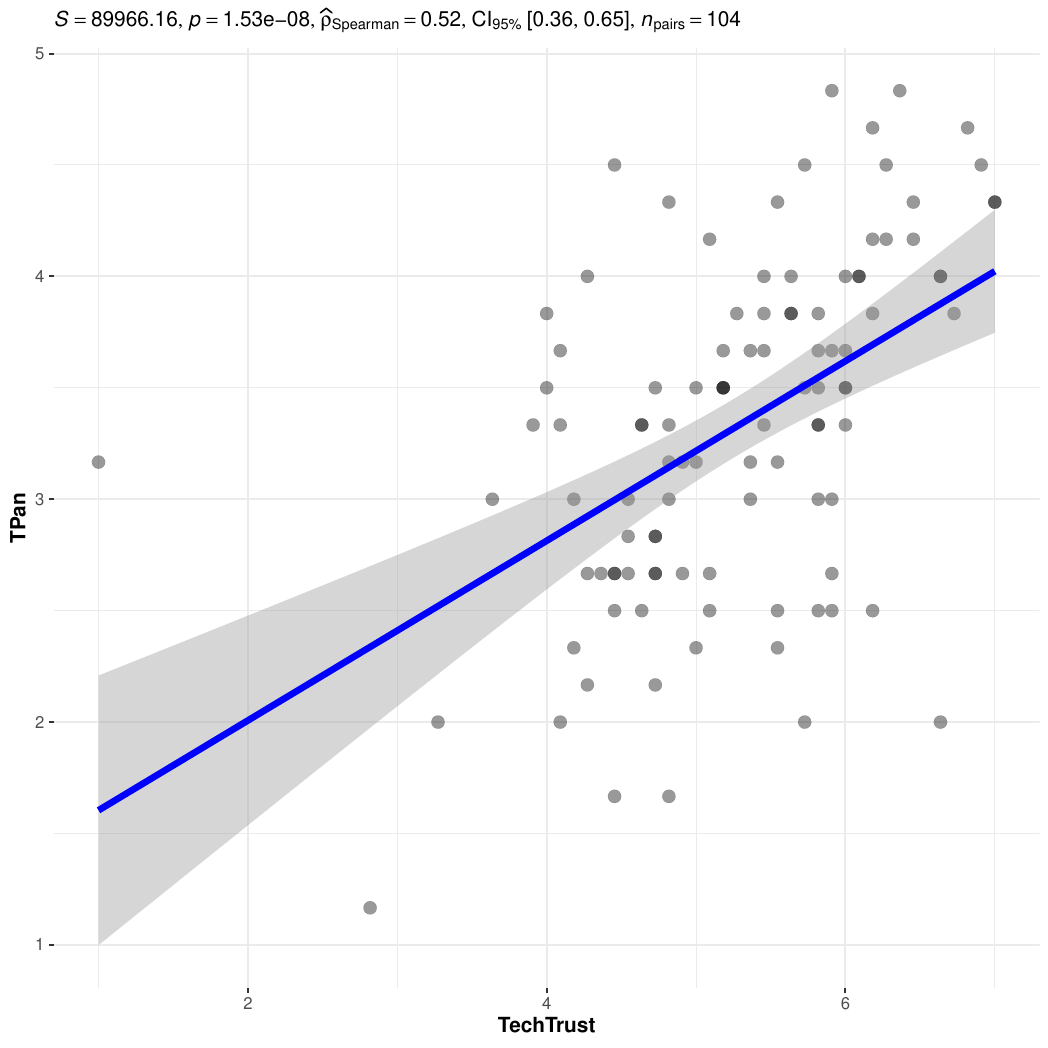}
\includegraphics[width=0.48\linewidth]{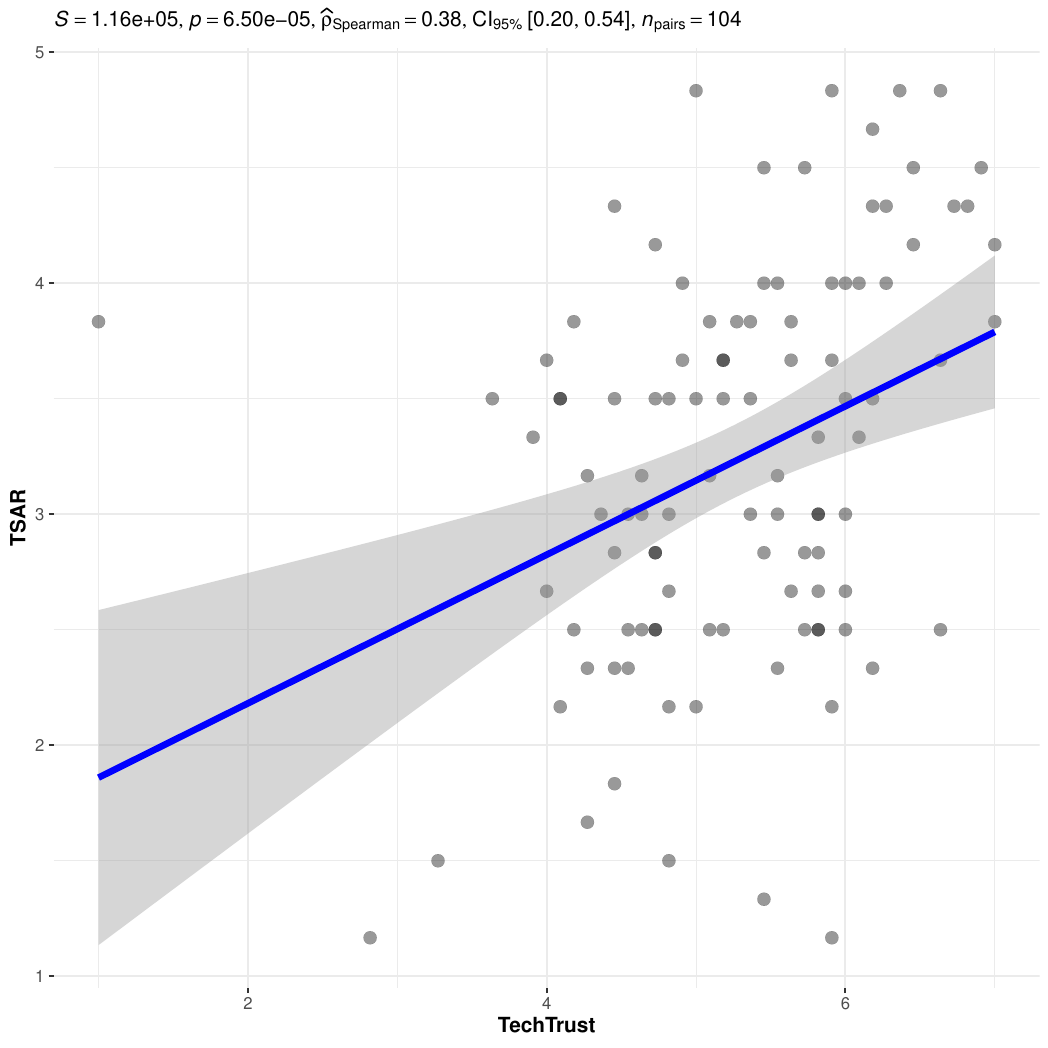}
\end{framed}
\caption{Comparison of Trust in technology against trust in pancake robot (left) and search-and-rescue (right)}\label{fig:h9both}
\end{figure*}

\newpage 

\section{Survey}\label{appx:survey}

The remainder of this document is the survey used. 

\includepdf[pages=-]{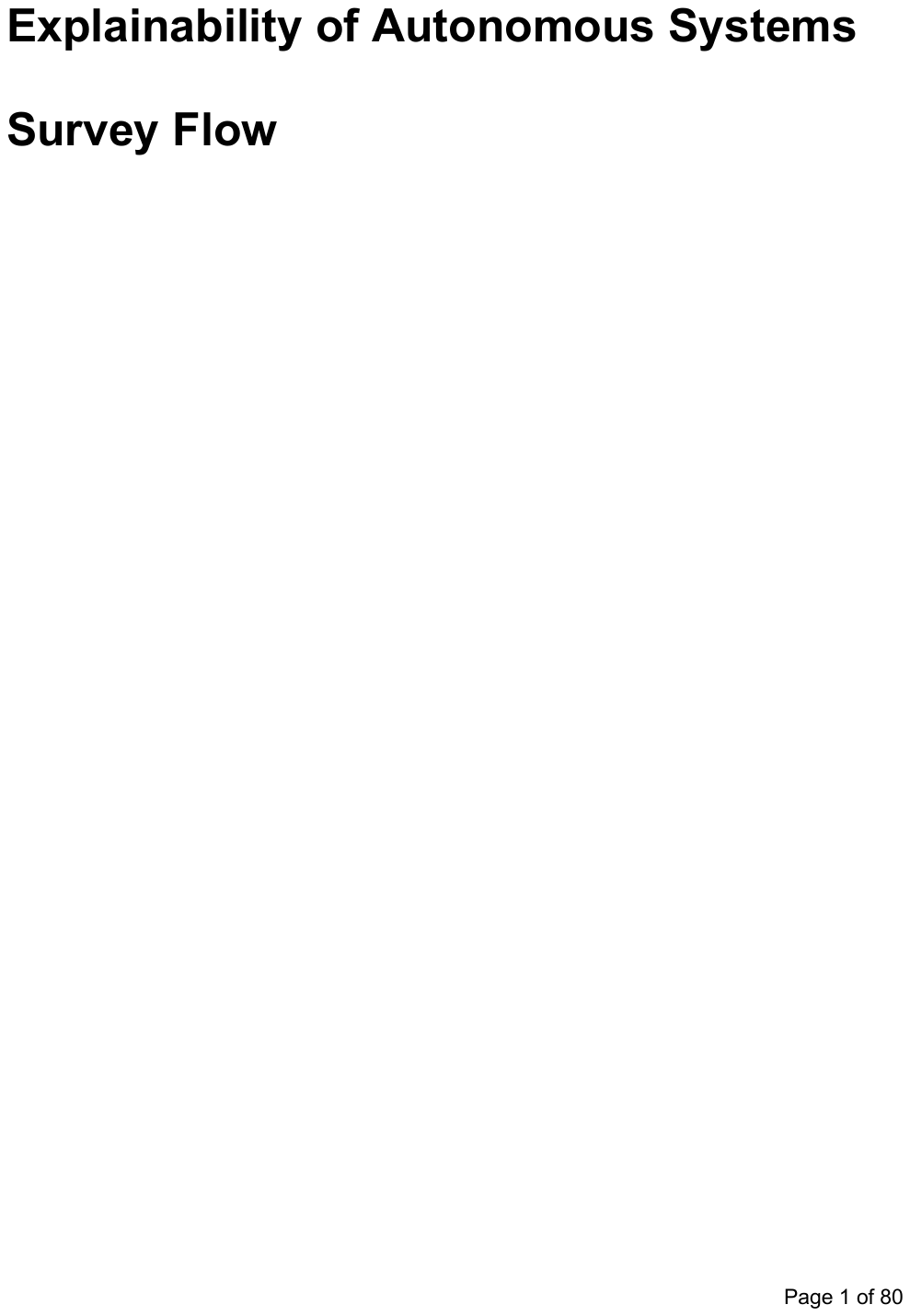}

\end{document}